\pdfoutput=1
\documentclass{article}
\usepackage{authblk}
\usepackage{graphicx,epsfig,amscd,amsfonts}
\usepackage{hyperref}
\usepackage{geometry}
\usepackage{xcolor}
\definecolor{customblue}{RGB}{0,81,158}
\usepackage{fancyhdr}
\usepackage{amsmath,amssymb,amsthm}
\usepackage{mathtools}
\usepackage{bm}
\usepackage{pgfplots}
\pgfplotsset{compat=newest}
\usepackage{tikz, listofitems}
\usetikzlibrary{calc, shapes, backgrounds, bending, decorations.pathmorphing, positioning, decorations.pathreplacing, arrows.meta}
\usepgflibrary{arrows}
\usepackage{subcaption}
\usepackage{enumitem}
\usepackage{adjustbox}
\usepackage{siunitx}
\usepackage{lineno}
\usepackage{float}
\usepackage{pifont}
\usepackage{stmaryrd}
\usepackage{wasysym}
\usepackage{marvosym}
\usepackage{subcaption}
\usepackage{stackengine}
\usepackage[T1]{fontenc}
\usepackage{yfonts}
\usepackage{textcomp}
\usepackage{lingmacros}
\numberwithin{equation}{section}

\begin{document}
\title{Developement of Reinforcement Learning based Optimisation Method for Side-Sill Design} 
\author[1]{Aditya Borse}
\author[1]{Rutwik Gulakala}
\author[1]{Marcus Stoffel\thanks{The corresponding author is M.~Stoffel. Email address: stoffel@iam.rwth-aachen.de}}
\affil[1]{Institute of General Mechanics (IAM), RWTH Aachen, \protect\\ Eilfschornsteinstraße 18, 52062 Aachen, Germany\vspace{3pt}}
\date{November 2024}
\maketitle

\begin{abstract}
Optimisation for crashworthiness is a critical part of the vehicle development process. Due to stringent regulations and increasing market demands, multiple factors must be considered within a limited timeframe. However, for optimal crashworthiness design, multiobjective optimisation is necessary, and for complex parts, multiple design parameters must be evaluated. This crashworthiness analysis requires computationally intensive finite element simulations. This challenge leads to the need for inverse multi-parameter multi-objective optimisation. This challenge leads to the need for multi-parameter, multi-objective inverse optimisation. This article investigates a machine learning-based method for this type of optimisation, focusing on the design optimisation of a multi-cell side sill to improve crashworthiness results. Furthermore, the optimiser is coupled with an FE solver to achieve improved results. 
\end{abstract}

\section{Introduction}
\label{sec:Intro}
Crashworthiness is an essential aspect of vehicle development. Designing for crashworthiness involves evaluating multiple designs with Finite Element (FE) simulations. These simulations are nonlinear and dynamic; therefore, they are computationally expensive, and specific knowledge and software are needed to perform them. Thus, evaluating multiple designs with such simulations is time-consuming and computationally costly \cite{Duddeck_08, Zhang_10}. Therefore, surrogate modelling and various optimisation methods are employed in the design process \cite{Schäffer_19, Zhang_10}. The car is getting more complex, and multiple objectives must be satisfied for optimal design \cite{Duddeck_08}. Often, these objectives are contradictory; therefore, the methods must find a compromise to get the optimal solution. However, due to increasing complexity, multiple parameters must be analysed to get the results, and therefore, there is a need for an efficient multi-objective optimisation method. 

Evolutionary computation algorithms are primarily used for optimisation. However, these can struggle with multi-objective optimisation involving multiple parameters \cite{Zhang_10}. For acceptable results, extensive population/ simulations are required; therefore, this article proposes the machine learning (ML)-based optimisation method. This approach extends the previous investigation of part optimisation using ML-framework \cite{Borse_AoM_2024}. In this framework, FE simulations are performed for data generation. It is essential to understand the critical parameters and crashworthiness objectives needed to be fulfilled by the part. Then, a regression-based FE surrogate is trained, and reinforcement learning (RL) is used to train an intelligent RL agent. This RL agent performs the parameter exploration and determines parameters to satisfy the objectives. 

Initially, a multi-cell side sill design is chosen for optimisation, and the optimisation problem is formulated. In the optimisation problem, the design variables are defined as optimisation parameters, while energy absorption and mass are considered as crashworthiness objectives. The FE simulations are carried out to train the regression-model-based surrogate. This surrogate model establishes the relationship between the parameters and objectives and reduces the computational cost during optimisation. Then, parameter exploration is used to satisfy the objectives through a custom RL environment. The validation is carried out by coupling the RL environment to the FE solver and optimising the design.

\section{Side sill design and optimsiation}
\label{sec:motivation}
Design for crashworthiness under side impact is challenging, as in a side collision, the vehicle structure has limited space to absorb energy and deform under impact. Also, the passenger safe space is very close to the side structures; therefore, much investigation is done to create the structures to handle side impacts effectively. One such component is the side sill, located along the vehicle's side. It helps absorb and redistribute the impact energy along with neighbouring components such as door reinforcement, cross-members and the B-pillar section \cite{Li_21, Xia_22}. The side sill is even more critical for electric vehicles, as most electric cars have the battery near the underside of the vehicle \cite{Belingardi_23, Nicoletti_21}. This area also housed the cooling components of the battery; therefore, the optimal design of the side sills is becoming increasingly important to ensure that the battery is unaffected in the crash \cite{Xia_22, Li_21}.

Numerous studies have investigated B-pillar, side sill and door reinforcements, as they are the first to come in contact with impacting bodies or poles in oblique pole impact testing \cite{Long_19, Wang_19, Belingardi_23}. However, side-sill design optimisation studies are inadequate \cite{Li_21, Xia_22}. Some literature refers to the side sill as a rocker. Generally, The side sill is a long beam with a square or rectangular cross-section. It can have a reinforcement structure in the cross-section. The reinforcement is also called inserts and can be made of reinforced plastic, aluminium or steel extrusions \cite{Belingardi_23, Li_21}, which results in the multi-cell side sill. 

The primary structural objective of the side sill design is to absorb the impact energy and redistribute the rest of the energy away from the passenger \cite{Li_21, Xia_22, Estrada_22}. The side sill must also be stiff to prevent excessive deformation under impact loading \cite{Xia_22, Li_24}. This stiffness also becomes one of the key objectives in designing the side sill. This approach can be dangerous and may cause injuries to the passengers. Therefore, dummy injury-based criteria must also be considered \cite{Horstemeyer_09}. Thus, the reaction forces and deceleration values under side impact can also optimised \cite{Li_24}. The side sill mainly absorbs the energy by collapsing but should not intrude into the safe space. Furthermore, due to various requirements, lightweight design is essential in vehicle development, and therefore, various composite and foam-filled materials are also investigated \cite{Djamaluddin_23}.

Multicell side sills are always investigated due to their energy absorption capabilities and strength under transverse impact loading \cite{Belingardi_23}. Therefore, a multicell side sill is designed and optimised in this article. 

Conventionally, evolutionary algorithms (EA), genetic algorithms (GA), and particle swarm optimisation (PSO) are used for optimisation in the automotive industry \cite{Duddeck_08, Shetty_15}. However, the results of optimisation methods depend on the user's knowledge and chosen hyperparameters \cite{Hayashi_22}. Recently, a combination of conventional optimisation has gained popularity. Still, the quality of the results again relies on the various function definitions. It needs multiple iterations with a large number of simulations to create populations and find the optima \cite{Duddeck_08, Li_20}. The Kriging, Response Surface Method (RSM) surrogate modelling and some optimisation are often used to reduce computational efforts \cite{Duddeck_08, Li_21, Li_24}. This is quite time-consuming, and evaluating all parts and load cases is not feasible.

Furthermore, the conventional optimisation methods fail to converge for more than three parameters and multi-objective optimisation (MOO) \cite{Li_20}. Further, if the initial optimisation problem is slightly changed, the conventional optimisation methods must be recomputed \cite{Wang_23}. Therefore, an ML-based optimisation process is developed in this article. The goal is to create a generalizable optimisation process that can be used for different problems with different initialisations without numerous changes. Deep reinforcement learning has recently been used for multiobjective optimisation problems \cite{Li_20, Zou_21, Doi_19}. It has been examined for crashworthiness design optimisation in this investigation.

In this investigation, the side sill is chosen as the part to be optimised for crashworthiness. This side sill is a multi-cell side sill with reinforcement and has various wall thicknesses. These wall thicknesses are optimised for maximum energy absorption while reducing the overall mass of the side sill. This article focuses on the optimisation method using machine learning for multiple parameters (wall thicknesses) and multi-objective (energy absorbed and mass) optimisation for side sill design. 

\section{Inverse optimisation function}
\label{sec:Optimisation}
This study investigates the optimal side sill's structural parameters that satisfy the predetermined crashworthiness objectives. Therefore, the inverse multi-objective optimisation problem needs to be solved. The optimisation method has to explore the parameter space and determine the optimal structural parameters that satisfy the objectives. Thus, the structural parameters can be defined as a parameter vector \(\boldsymbol{T}\).

\begin{equation}
\label{Input_variables}
    \begin{array}{c}
        \quad \boldsymbol{T} = \left[t_1~t_2~ t_3...\right]^T \\
    \end{array}
\end{equation}

Where the structural parameter vector \(\boldsymbol{T}\) consists of \(t_i\), which is the wall thicknesses of the side sill and can have physical constraints on how much they can change. It depends on the design and manufacturing constraints. Accordingly, the inverse multi-objective optimisation problem can be defined by Eq. \ref{Opt_problem}, and each crashworthiness objective can be defined as a function of input vector \((f_\beta(\boldsymbol{T}))\). Here, \(\beta\) can take two values [1,2] for two objectives.

This approach is based on the research associated with multi-objective optimisation \cite{Lim_2020_spaceframe}. The optimisation function (\(O(\boldsymbol{T})\)) can be described as the combination of the functions of crashworthiness objectives. Here, the optimisation function (\(O(\boldsymbol{T})\)) is defined as the weighted sum of the difference between the current value of energy objectives (\(f_1(\boldsymbol{T})\)) and user-defined crashworthiness metrics (\(M_\text{1}(\boldsymbol{T})\)) and mass of the side sill as objective (\(f_2(\boldsymbol{T})\)). The first component is to maximise the energy absorption, and the second part is to minimise the mass of the side sill as much as possible. Users can vary the weight depending on the importance of objectives and create a single optimisation function from multiple objectives. 

\begin{equation}
\label{Opt_problem}
    \begin{array}{rl}
        \min &\quad O(\boldsymbol{T}) = \displaystyle\min_{\boldsymbol{T}} w_1*(f_1(\boldsymbol{T})-M_1(\boldsymbol{T})) + w_2*f_2(\boldsymbol{T}) \\
        \text{Subject to:} &\quad  \text{Physical constraints on values of } \boldsymbol{T} \\
    \end{array}
\end{equation} 

Here, the energy absorption is to maximise to be equal or greater than ideal user-defined crashworthiness metrics (\(M_\text{1}\)), and the mass has to be minimised as much as possible; therefore \(w_1\) is considered as 1 and \(w_2\) as -0.5. This gives more importance to energy absorption than the mass objective. Thus, the overall objective becomes to minimise the difference between energy objectives and reduce the mass of the side sill.

\section{Method}
\label{sec:method}

\subsection{Finite Element impact simulations}
\label{sec:FE_simulations}
\begin{figure}[H]
    \centering
    \includegraphics[width=0.5\linewidth]{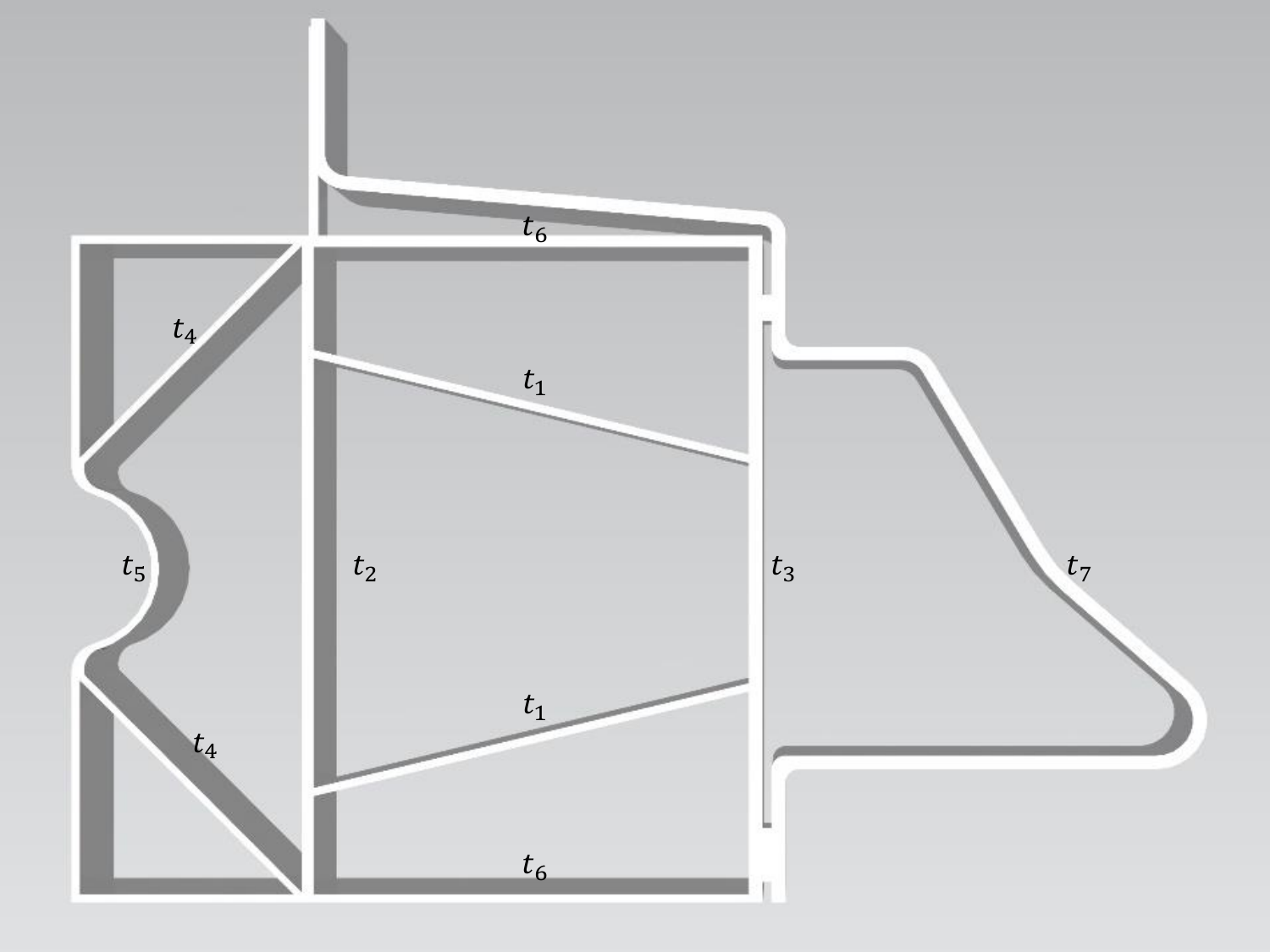}
    \caption{Cross-section of the side sill with the varied wall thickness}
    \label{fig:side_sill_section}
\end{figure}

The side sill is essential in a side impact crash; therefore, this article evaluates multi-cell side sill design and optimises the design parameters to fulfil the crashworthiness objectives. The side sill is designed for side crashes and redistributes energy away from the passenger safe space along with the neighbouring B-pillar and door reinforcement parts. 

The design of the side sill is shown in Figure \ref{fig:side_sill_section}. Here, the side sill comprises two parts, the inner sill section and the front part, which are welded as shown in Figure \ref{fig:side_sill_section}. The inner sill is a rectangular section with reinforcement flanges. The front part is the protruded part on the right side, which helps quickly absorb the impact energy in the initial stages. The inner sill section and the front part have different wall thicknesses in various areas, as illustrated in the figure. The article refers to the inner sill and the front part as the side sill. 

\begin{figure}[H]
    \centering
    \includegraphics[width=0.5\linewidth]{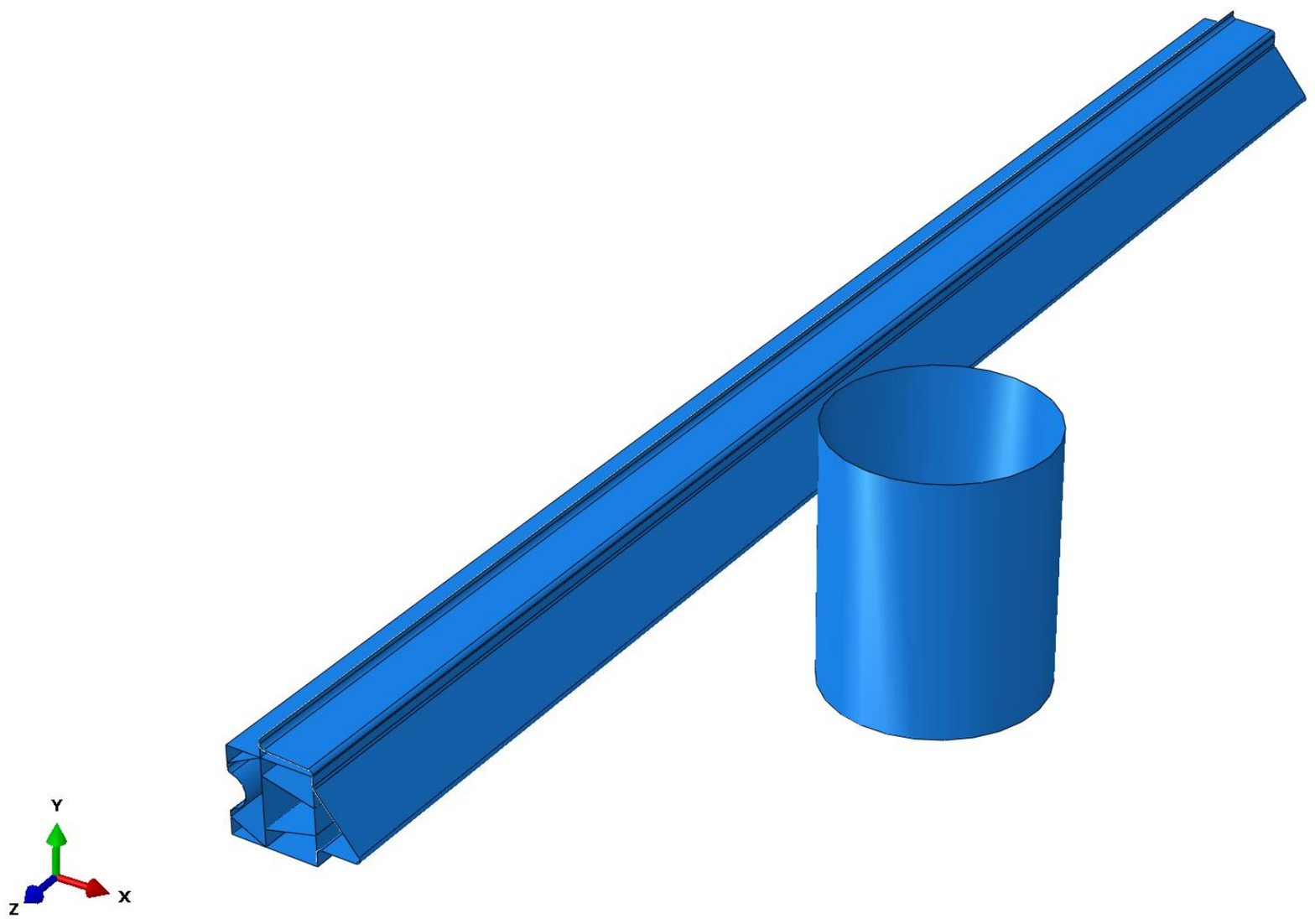}
    \caption{Assmebly view of one of the models of side sill and oblique pole impact}
    \label{fig:assembly}
\end{figure}

The side impact is simulated to be similar to the EuroNCAP oblique pole side impact testing protocol. In this testing, the test vehicle is accelerated at the fixed pole. However, in the FE simulations carried out in this investigation, the side sill is stationary, and the pole is impacted with an initial velocity of 8 m/s, which is approximately 28.8 km/h. The angle of the impact is 75 degrees. The pole is modelled as a rigid cylindrical structure with a diameter of 250 mm, a height of 300 mm, and a mass of 50 kg. 

\begin{table}[]
    \centering
    \caption{Wall thickness variation range and the step size}
    \label{tab:thickness_variation}
    \begin{tabular}{@{}cccc@{}}
        \hline
        thickness & min value & max value & step size \\
        \hline
        \(t_1\) & 1.5 & 3.0 & 0.1 \\
        \(t_2\) & 2.0 & 4.0 & 0.2 \\
        \(t_3\) & 2.0 & 4.0 & 0.2 \\
        \(t_4\) & 1.0 & 3.0 & 0.1 \\
        \(t_5\) & 1.5 & 3.5 & 0.1 \\
        \(t_6\) & 2.0 & 4.0 & 0.2 \\
        \(t_7\) & 2.0 & 4.0 & 0.2 \\
        \hline
    \end{tabular}
\end{table}

The complete side sill is modelled with a shell as an extrusion of length 2 m in Simulia's ABAQUS preprocessor (version 2021) \cite{Abaqus_2009}. The wall thicknesses of the complete side-sill are varied within the range shown in Table \ref{tab:thickness_variation} to generate various designs. Each simulation's thickness value is chosen randomly within the range, and the side impact pole simulation is carried out. Then, the simulation output files are saved. From these output files, the objective values are obtained from the reaction forces, energy and mass values.

\begin{table}[!htb]
\caption{Material properties}
\begin{minipage}{.5\linewidth}
      \subcaption{Elastic properties}
      \label{tab:Elastic_properties}
      \centering
        \begin{tabular}{@{}llllll@{}}
        \hline
        \textbf{Density} & \textbf{Young's modulus} & \textbf{Poisson's ratio} \\
        \si{\kg/\cubic\m} & \si{\giga\pascal} & - \\
            \hline
            2700 & 70 & 0.33 \\
            \hline
        \end{tabular}
    \end{minipage}
    \begin{minipage}{.5\linewidth}
      \centering
        \subcaption{Plasticity properties}
        \label{tab:Plastic_properties}
        \begin{tabular}{@{}llllll@{}}
            \hline
            \textbf{Yield stress} & \textbf{Plastic strain} \\
            \si{\kg/\square\m} & - \\
            \hline
            80 & 0 \\
            115.0 & 0.024\\
            139.0 & 0.049 \\
            150.0 & 0.079 \\
            158.0 & 0.099 \\ 
            167.0 & 0.124 \\
            171.0 & 0.149\\ 
            173.0 & 0.174 \\
            \hline
        \end{tabular}
    \end{minipage} 
\end{table}

The material is assumed to be bilinear elastoplastic material, and its properties are shown in Table \ref{tab:Elastic_properties} and \ref{tab:Plastic_properties}. The weld radius is 1.5 mm, and the weld points are 100 mm apart, including the endpoints. The complete side sill has meshed with S4R elements of 10 mm size and the pole with R3D4 elements of 5 mm mesh size. The interaction is considered general frictional with a frictional coefficient of 0.2 along the tangential direction, and the endpoints of the side sill are considered rigidly fixed to the neighbouring parts. The simulation is solved using ABAQUS's explicit dynamic solver with a geometrical nonlinearity of 0.1 seconds. 

A total of 310 FE simulations are completed. From these simulations, the peak contact force (\(PCF\)), energy (\(EA_\text{ss}\) and \(EA_\text{f}\)), and mass of the complete sill (\(Mass\)) are extracted. From this FE simulation output, the energy absorbed by both the parts of the side sill and the mass of the side sill are selected as the crashworthiness objectives for the inverse optimisation problem. . These objectives and the corresponding wall thicknesses form the simulation database and are further used to train the ML-based regression surrogate model.

\subsection{ML-based regression surrogate}
\label{sec:ML_regression}
In optimisation, multiple designs must be evaluated to understand the relation between the structural parameter and corresponding crashworthiness objectives. Therefore, a regression model is trained as the FE surrogate model, as the FE simulations are time-consuming. The regression model is crucial in the optimisation process. Therefore, an accurate and time-efficient regression model is needed. However, the data is analysed to train the correct regression model, and then the input-output for the regression model is chosen. 

\begin{figure}
    \centering
    \includegraphics[width=0.5\linewidth]{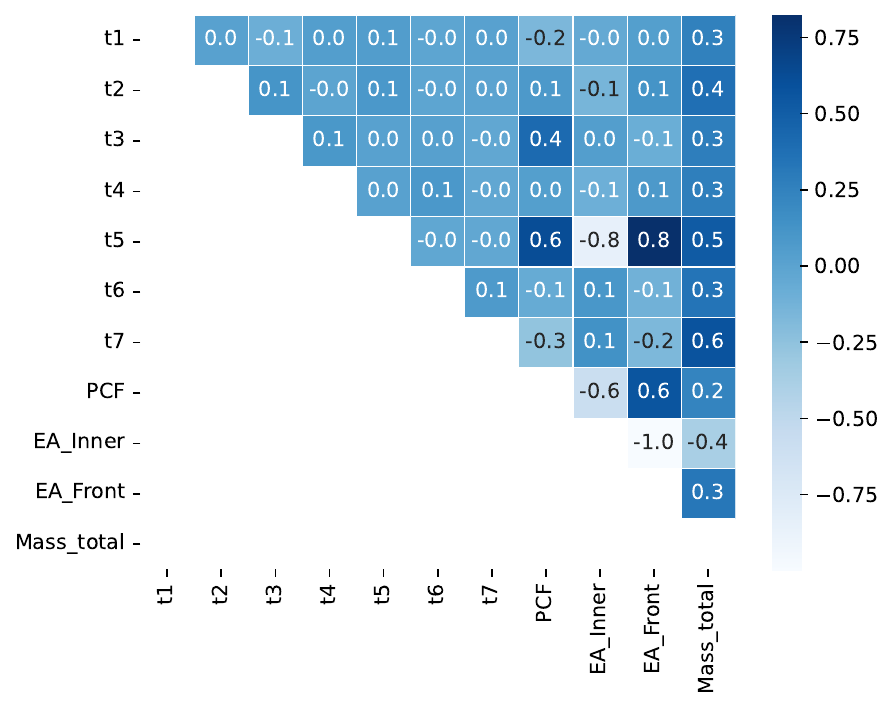}
    \caption{Data-correlation matrix of the FE simulation database}
    \label{fig:data_analysis}
\end{figure}

The data correlation matrix is shown in Figure \ref{fig:data_analysis}. The energy absorbed by the inner sill (\(EA_\text{ss}\)) and the front part (\(EA_\text{f}\)) and the total mass of the side sill (\(Mass\)) is chosen as the output of the regression model and the seven wall thicknesses as the input to the regression model. The regression model is Keras sequential \cite{chollet2015keras} with three hidden layers and is trained with Adam optimiser \cite{Adam_arXiv}. The hyperparameters, hidden neurons, and the learning rate are chosen using the Keras tuner with a maximum trial of 10 and two executions per trial to get the optimal hyperparameters. The hyperparameter range is listed in Table \ref{tab:hyperparameter_variation}.

\begin{table}[]
    \centering
    \caption{Range of the hyperparameters for hyperparameter optimisation}
    \label{tab:hyperparameter_variation}
    \begin{tabular}{@{}cccc@{}}
        \hline
        Hyperparameters & Minimum value & Maximum value & Sampling method \\
        \hline
        Hidden neuron in hidden layer 1 & 32 & 512 & Linear \\
        Hidden neuron in hidden layer 2 & 32 & 512 & Linear \\
        Hidden neuron in hidden layer 3 & 32 & 512 & Linear \\
        Learning rate for Adam optimiser & 0.0001 & 0.01 & Log \\
        \hline
    \end{tabular}
\end{table}

The FE simulation data is divided into two parts: the training dataset and the testing dataset. Using the Keras library \cite{chollet2015keras}, the split is 0.8 and 0.2. The output has varied scales; therefore, the outputs are standardized using StandardScaler from Sci-kit learn library \cite{scikit-learn}. The training results are discussed in the section \ref{sec:ML_regression_results}.

\subsection{Optimisation using Genetic algorithm}
\label{other_Optimisation_methods}
The genetic algorithm (GA) is a method for solving both constrained and unconstrained optimisation problems based on natural selection, which drives biological evolution. It repeatedly modifies a population of individual solutions. At each step, the genetic algorithm selects individuals from the current population to be parents and uses them to produce the children for the next generation. Over successive generations, the population evolves toward an optimal solution. The key elements of GA are chromosome representation, selection, crossover, mutation, and fitness function computation. Initially, a population $P$ of $n$ chromosomes are randomly selected. Two chromosomes say $C_1$ and $C_2$, are selected from the population $P$ according to the fitness value. The single-point crossover operator with crossover probability ($C_p$) is applied on $C_1$ and $C_2$ to produce an offspring, say $O$. Thereafter, a uniform mutation operator is applied to produced offspring ($O$) with mutation probability ($M_p$) to generate $O$. The new offspring $O$ is placed in a new population. The selection, crossover, and mutation operations will be repeated on the current population until the new population is complete. 

The current study replaces finite element simulation as the objective function with a fully connected network. The network is trained on design parameters as input, while energy is absorbed and mass as output. 

\subsection{Optimisation using Network inversion}
The network inversion method using a multilayer neural network was proposed to solve the optimisation process as an inverse problem. The goal of a neural network in supervised learning is to achieve the minimum possible error for an input $x$ with respect to a target $Y$, and this error can be mathematically expressed as

\begin{equation}
    E = |F(x) - Y| < \tau
\end{equation}

Where t is the allowable tolerance level for the deviation of the neural network output $F(x)$ from the desired target $T$. The mapping $F(x)$ describes the result of the ANN for the given input vector $x$. 

While a neural network during forward mapping calculates an output for a given input pattern, the inversion of a neural network involves the calculation of input patterns that would lead to the desired output. This can be realized by implementing
the gradient descent in the input space, wherein a set of input vectors $X$ are calculated such that the network error is minimized. The calculation of these input vectors is realized as

\begin{equation}
    x^{l+1} = x^l - \eta \frac{\partial E}{\partial x^l}
\end{equation}

where $\eta$ denotes the learning rate, $x^0$ is the initial guess for
the input vector, the index $l$ refers to the current input vector
, and the index $(l+1)$ to the updated input vector. In the current study, for a fully connected feed-forward network (FCFFN), the backpropagation algorithm is used for training \cite{kindermann}. Therefore, the fully connected network described in the previous sub-section, which is adapted as the objective function, is used as the network to be inverted ie forward network. The inverse network is then trained using the loss of the forward network. However, one of the major problems in inverting an ANN is its uniqueness. Due to the fact that there exist several input patterns for a single output of fractal dimension, it would be difficult to generate the original model parameters through the ANN inversion. To this end, constraints were added in the form of boundaries for input variables, so that they are not exceeded during prediction.

\subsection{Optimisation using Reinforcement Learning}
\label{sec:RL}
Reinforcement learning (RL) has become popular in the last decade due to its ability to learn from experiences similar to humans. Therefore, reinforcement learning is being used in various domains. This article uses reinforcement learning to explore wall thickness parameters and obtain optimal parameter values for the side sill design. Two objectives define the optimality of the parameters: energy absorbed by the side sill and the mass of the side sill. The parameters are searched to maximise the total energy absorption while reducing the mass of the side sill. 

Therefore, a custom RL environment is created. RL environment can be understood as the virtual world where RL agents can take some action, thereby changing the state of the world. Depending on whether the action was beneficial to the virtual world or not, the world gives the RL agent feedback, and the RL agent learns from this feedback to take actions that are useful to the world. In this investigation, the RL environment provides the seven wall thickness values as the state variables. It also takes an external array having energy absorbed by the inner sill (\(EA_\text{ss}\)) and the front part (\(EA_\text{f}\)) and the total mass of the side sill (\(Mass\)) to get the ideal combined energy required to be absorbed by the side sill and \(Mass\) is just information parameter. 

The RL environment allows the RL agent to change one of the thicknesses in a predefined step and take one action at a time. The action the RL agent can take on each thickness is to either increase or decrease the thickness value by a given step size, and these step sizes for all thicknesses are identical to the step size shown in Table \ref{tab:thickness_variation}. The RL agent can vary the thickness in the given range only. The RL agent takes random action at first to gain experience, and a numerical reward is given to the RL agent for every action. From this reward, RL agents learn the dynamics of change in the state variables and, according, rewards at every step. 

The agent can only understand the RL environment through rewards, so the appropriate definition of the reward function is essential. The reward function defined in the custom RL environment is based on the difference between the current energy absorbed and the ideal combined energy from the external array and the current mass of the side sill. Therefore, to get the current values of \(EA_\text{ss}\), \(EA_\text{f}\) and \(Mass\) from the current state of the RL environment, the trained regression model is used. The scales of both the objectives, combined energy and mass of the side sill, are different; therefore, both objectives are scaled within [1,100] before calculating the reward function. This would help keep the reward unbiased to higher terms, such as combined energy here. 

The current energy absorbed value (\(f_1(\boldsymbol{T})\)) is defined as the sum of the scaled current \(\overline{EA_\text{ss}}\) and \(\overline{EA_\text{f}}\) and ideal energy absorbed value (\(M_\text{1}(\boldsymbol{T})\)) as the sum of the scaled ideal \(EA_\text{ss}\) and \(EA_\text{f}\) from the external array. The mass objective (\(f_2(\boldsymbol{T})\)) is defined as the scaled current mass of the side sill \(\overline{Mass}\). The overline values are for the current state of the environment and can change with the RL agent's action. Then, the reward is defined as follows:

\begin{equation}
\label{Eq:Reward_fun}
    \begin{array}{c}
        \quad \boldsymbol{R} = (f_1(\boldsymbol{T}) - M_\text{1}(\boldsymbol{T})) -  0.5*f_2(\boldsymbol{T})\\
    \end{array}
\end{equation}

Furthermore, two termination conditions are defined to avoid the infinite loop while training. The first termination condition (T1) is to train the model for 500 steps in every episode. The second condition (T2) terminates the training if the mean sum of the absolute difference between the scaled current and ideal energy absorption value is less than 5. If the termination happens by the second condition +10, an additional reward is given to the RL agent to drive it to reach optimal values as soon as possible. The second condition ensures that the RL agent can settle on the optimal thickness value even if the user-defined ideal values are incorrect, as some objectives might be impossible to achieve for any selection of parameters. 

As the RL environment is model-free and the RL agent can only take discrete action on the thickness values, a limited number of RL agents can be used. Therefore, the stable-baselines-3 (SB3) library \cite{stable-baselines3} is used for training as it provides several agents for our environment. However, only Advantage Actor-Critic (A2C) is trained and evaluated for this investigation. The RL agent is the default A2C agent with Adam Optimiser, but it is only trained for a maximum of 200 episodes. The ideal energy-mass array [800, 600, 13] is provided as the input of the ideal objective array for training. The agent is saved at the end of the training. This saved agent can help evaluate the ideal thickness values for different objectives with the same RL environment. However, in the FE simulation database, most of the ideal combined energy absorption is around 1400 J, and even the trained agent cannot estimate better energy absorption values while reducing the mass. This is due to the limitation of the extrapolation capabilities of the regression model, further enforced by the limited database.

Therefore, the RL environment is coupled with the Abaqus for further evaluation. The environment is slightly changed to feed the state variable (current thickness values) as the input to the Abaqus to model the similar side pole impact and run the dynamic simulation. After the completion of the Abaqus dynamic simulation, the current energy and mass values, \(EA_\text{ss}\), \(EA_\text{f}\) and \(Mass\), are fed to the RL environment and the reward is calculated for the RL agent's action using same reward function as mention above \ref{Eq:Reward_fun}. Furthermore, the first termination condition is changed to 20 as the maximum number of steps, while the second termination condition is still the same. As the simulations are run in the FE solver, the simulations take longer time than the regression model to get the objective values for the current thicknesses. Therefore, the T1 is changed to 20 maximum number of steps.

To estimate the ideal parameters for maximum energy absorption while minimising the side sill mass, a higher ideal energy-mass array [825, 625, 14] is provided as the ideal objective array for the coupled RL environment. Only one episode is run while evaluating the optimisation results using the coupled RL and trained A2C agent. The details about the results of these optimisation methods are described in Section \ref{sec:Optimisation_results}.

\section{Results}
\label{sec:results}

\subsection{FE simulations results}
\label{sec:FE_simulations_results}
The FE simulation results are generally in the form of the various output variables extracted at every time step. These results cannot be directly used in the optimisation process. Therefore, the required objectives and the corresponding structural parameters are extracted from the simulation results. This forms the FE simulation database and can be used to train the regression model. 

The structural parameters, the seven wall thickness values of the side sill section and objectives, peak contact force (\(PCF\)), total energy absorbed by the inner sill section (\(EA_\text{ss}\)), energy absorbed by the front part (\(EA_\text{f}\)) and mass of the complete side sill (\(Mass\)) are extracted to form the FE simulation database. The simulation database is generated using a Python script to read the simulation output files and obtain the required parameters and objective values. As mentioned, the thickness values vary randomly to create multiple side sill designs, which are then simulated for the same impact conditions. The average time for the completion of the simulations is around 10 minutes. In the FE simulation database, the average total energy absorbed and mass of the side sill is 1400 J and 14.5 kg, respectively. 

\subsection{ML-based regression model outcomes}
\label{sec:ML_regression_results}
As mentioned above, the regression model was trained on the FE simulation database. In the optimisation problem, energy absorbed by the inner sill section (\(EA_\text{ss}\)), energy absorbed by the front part (\(EA_\text{f}\)) and mass of the complete side sill (\(Mass\)) are considered as the output objectives and the seven wall thickness values as the input to the regression model. The total energy absorbed by the side sill is sum \(EA_\text{ss}\) and \(EA_\text{f}\). An appropriate choice of hyperparameters for the regression model is required for good results; therefore, hyperparameter tuning is carried out using a Keras tuner. The Keras tuner is set up to minimise the validation mean absolute error at the end of the training. The training and the hyperparameter tuning took  3 min 19 sec. The test dataset's validation mean absolute error (MAE) and mean square error (MSE) are 0.0672 and 0.0079, respectively.

\begin{figure}
    \centering
    \includegraphics[width=0.5\linewidth]{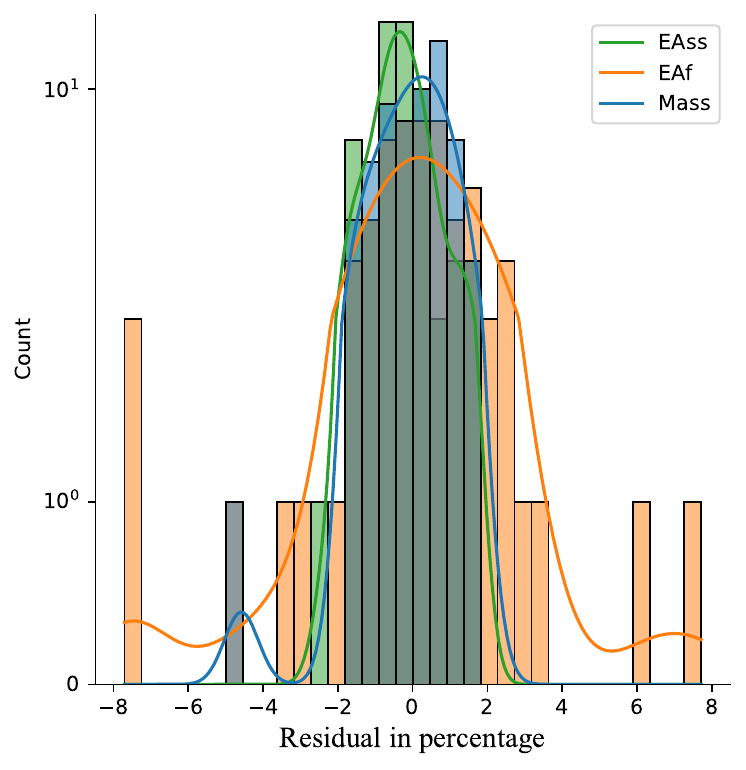}
    \caption{Residual distribution for the validation of the trained regression model}
    \label{fig:res_analysis}
\end{figure}

This FE surrogate is critical for the proper training of the RL agents; therefore, the accuracy of the trained surrogate is vital. The trained model is further evaluated with the residual distribution of the prediction of the test dataset and is shown in Figure \ref{fig:res_analysis}. The residual is the relative percentage difference between the true and the predicted value with respect to the true value. The resulting residual values have a normal distribution, and the values are within +/- 5\%. Thus, the regression model is accurate enough to be used as an FE surrogate in the RL environment for training the RL agents. 

\subsection{Optimisation results}
\label{sec:Optimisation_results}
\begin{figure}
    \centering
    \includegraphics[width=0.5\linewidth]{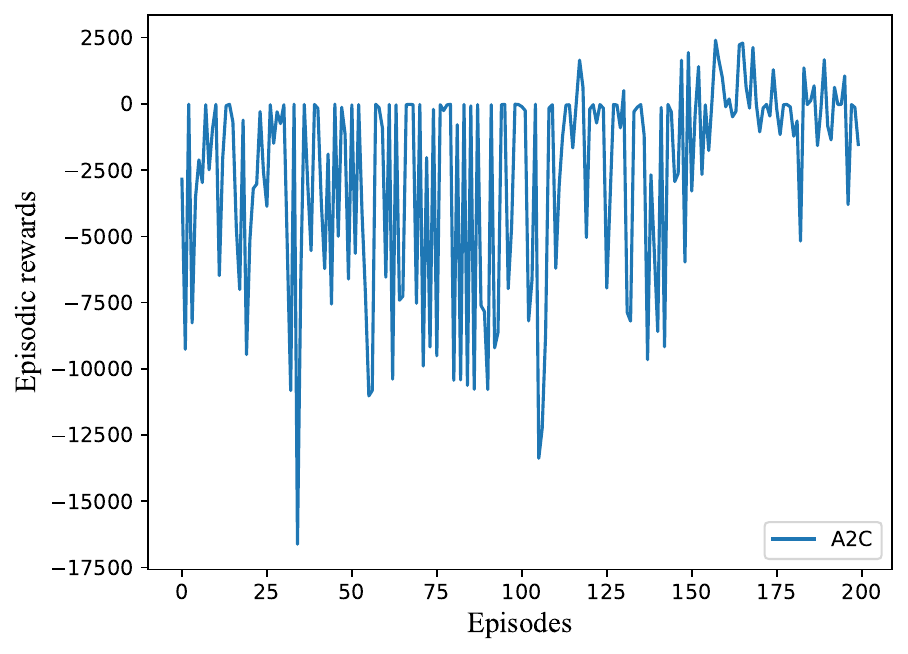}
    \caption{Training of SB3- A2C agent- Episodic rewards over episode}
    \label{fig:Episodic_reward_a2C}
\end{figure}

For the initial training of the SB3 A2C agent, the ideal objective array given is [800, 600, 13], and the training took 31.1746 min. A logger wrapper is used to record the training of the A2C agent and the episodic reward over the episode plot is shown in Figure \ref{fig:Episodic_reward_a2C}. As requested, the training terminates after 200 episodes using the callback during training. Initially, the agent chooses the actions at random, and thus, while gaining experience, fluctuating rewards are obtained, but after 150 episodes, the agent seems to take appropriate actions. Therefore, at the end of the training, some positive rewards are generated, and the reward fluctuations are in a limited region. This is due to further exploration of parameters. 

The agent is saved after a maximum of 200 episodes using the callback function. The saved agent is loaded to determine the optimal thickness value if the ideal objective array given is [900, 600, 13]. The results are [2.8, 4.0, 2.2, 3.0, 3.3, 2.1, 2.1] as the thickness values and corresponding combined energy absorption is 1403.37 KJ with mass as 14.71 kg. The results are good but not optimal, but it is probably due to the expected ideal energy absorption of 1500 kJ, which is not observed in the FE simulation database and is far away from the average 1400 J observed in the FE simulation database. This higher expected value can lead to extrapolation issues with the regression model. As seen in the FE database, it is possibly impossible due to physical constraints on the thickness value.

\begin{table}[]
    \centering
    \caption{Optimisation results of different methods}
    \label{tab:Optimisation_results}
    \begin{tabular}{@{}cccc@{}}
        \hline
        Optimisation method & thickness values (t1-t7) & Total energy absorbed & Mass\\
        - & [mm] & [J] & [kg]\\
        \hline
        Genetic Algorithm     & 2.1-3.2-3.6-2.9-1.5-3.6-3.8 & 1370.35 & 15.59 \\
        Network Inversion     & 2.0-2.7-2.7-2.0-2.3-2.9-2.6 & 1412.95 & 13.16 \\
        RL with Abaqus (T2=3) & 1.7-2.0-2.0-1.0-3.0-3.4-4.0 & 1417.36 & 14.49 \\
        RL with Abaqus (T2=4) & 1.7-2.0-2.0-1.0-3.0-3.4-4.0 & 1417.36 & 14.49 \\
        RL with Abaqus (T2=5) & 1.7-2.0-2.0-1.0-3.0-3.4-4.0 & 1417.36 & 14.49 \\
        \hline
    \end{tabular}
\end{table}

Therefore, the RL environment is directly coupled to the Abaqus for better results. The coupled RL environment with an ideal objective array as [825, 625, 14] is trained for a maximum of 20 steps and a single episode. The expected ideal energy absorption is given as 1450, higher than the FE simulations database. The coupled RL and Abaqus optimisation took 53.85 min. 

\begin{figure}
     \centering
     \begin{subfigure}[b]{0.48\textwidth}
        \centering
        \includegraphics[width=\textwidth]{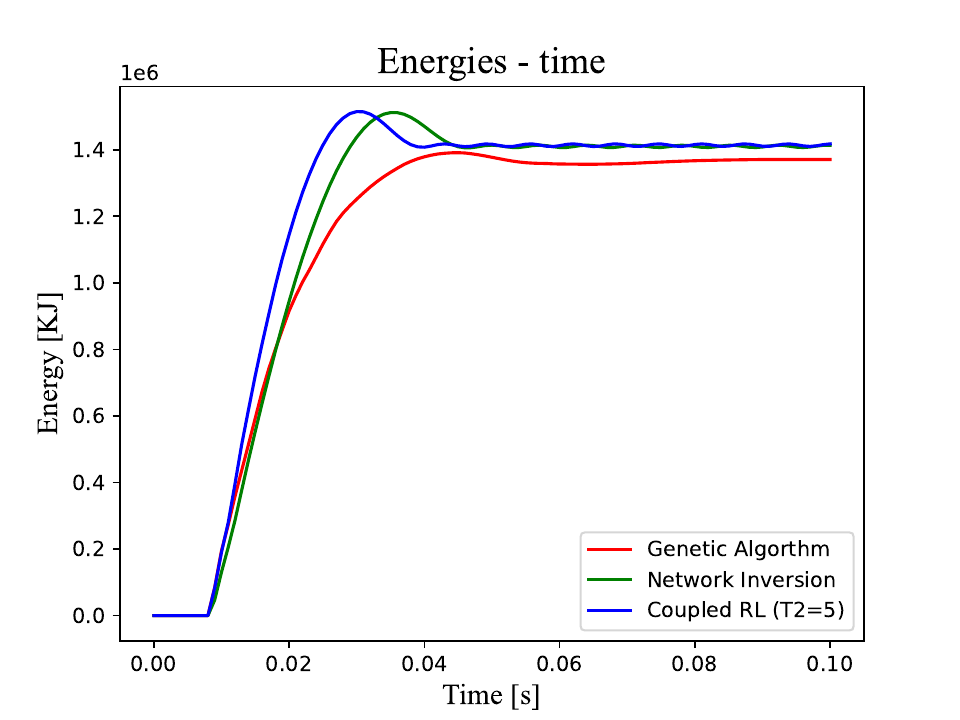}
        \caption{Comparison of energy absorption of different optimisation methods using FE simulation validation}
        \label{fig:Energy_comparison}
     \end{subfigure}
     \hfill
     \begin{subfigure}[b]{0.48\textwidth}
        \centering
        \includegraphics[width=\textwidth]{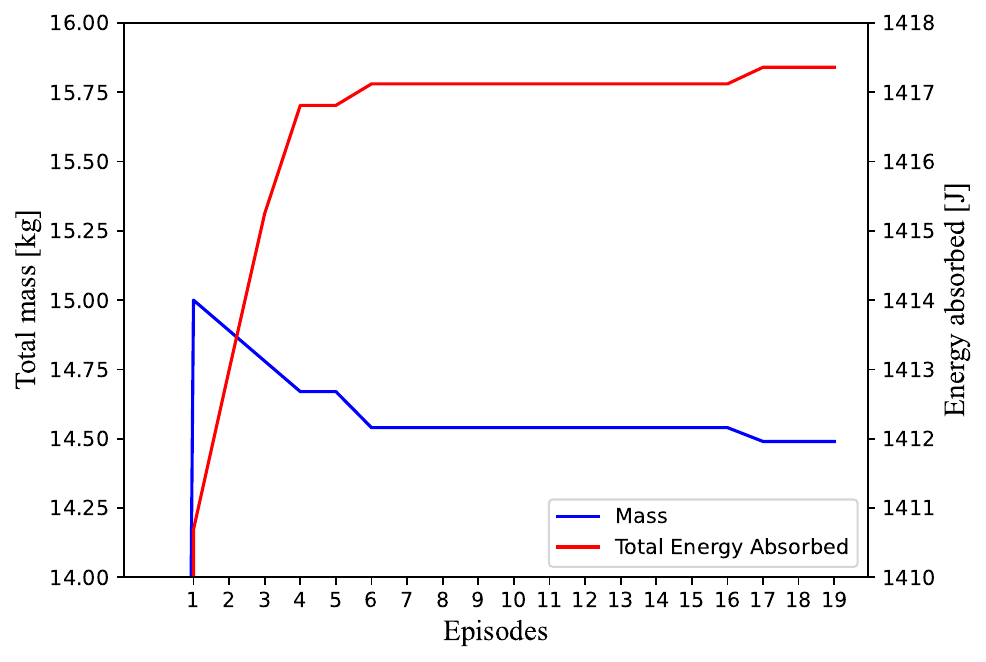}
        \caption{Validation of RL environment coupled to Abaqus (T2=5): Energy absorbed and mass of complete side sill over episodes}
        \label{fig:Cuoupled_rl_tol5}
     \end{subfigure}
     \hfill
        \caption{Results of different optimisation methods and evaluation of coupled RL environment}
        \label{fig:three graphs}
\end{figure}

To compare the different optimisation methods, FE simulations are run with the obtained thickness values and the total energy absorbed and mass of the side sill is obtained. The results of the same RL environment coupled with the Abaqus with different T2 conditions along with the Genetic Algorithm (GA) and Network Inversion (NI) are shown in Table \ref{tab:Optimisation_results}.  The genetic algorithm's results are not satisfactory compared to other methods. The absorbed energy results of NI and the coupled RL are comparable, as shown in Figure \ref{fig:Energy_comparison}. However, as in the optimisation problem, more importance is given to the total energy absorbed over the side sill's mass; therefore, the coupled RL shows more energy absorption with a slightly increased mass. 

The results from the coupled RL environment are acceptable, and further validation using the FE environment is not needed. However, the thickness values obtained from the coupled RL environment are the same even for different termination conditions. The coupled RL environment also took 20 steps while optimising, as shown in Figure \ref{fig:Cuoupled_rl_tol5} for coupled RL environment with T2=5, incrementally improving the results. From the results, it could be possible to use similar coupled RL environments for other optimisation problems. 

Thus, further exploration of RL-based optimization could be valuable. The training of the RL agent relies on the reward function within the RL environment. Therefore, different reward functions could be examined for efficient agent's training. Additionally, in this investigation, only A2C is utilised, but other suitable agents for a model-free RL environment could be reviewed. It may be possible to achieve improved results with a suitable agent and reward function. 

\section{Conclusion}
\label{sec:conclusion}
This article investigates a machine learning-based optimisation process for inverse multiparameter multi-objective problems. It examines the multi-cell side sill with reinforcement and focuses on optimising the wall thickness. The crashworthiness objectives are to maximise the energy absorption while reducing the side sill's mass. Therefore, the seven wall thickness values need to be determined to satisfy the objectives. Thus, the inverse multi-objective optimisation problem needs to be solved.

The FE simulation of the side sill is initially performed based on the oblique side pole impact tests, creating the FE simulation database. A machine learning-based regression model is trained to serve as a surrogate for the FE simulations from this database. Furthermore, Reinforcement learning (RL) is utilised for parameter exploration and determining the optimal parameters. A custom RL environment is designed where an intelligent RL agent can change the thickness parameters to attempt to satisfy the objectives. Furthermore, the RL environment is integrated with Abaqus, allowing the trained RL agent to evaluate and optimize the side-sill design to achieve better results than using only the regression as an FE surrogate. 

The RL environment, coupled with the Abaqus solver, provided encouraging results for the current multi-parameter, multi-objective problem. For similar optimisation problems, the method could potentially benefit from automating the optimisation process by coupling it to the system. The optimisation process with RL is purely based on the mathematical objective function; theoretically, it could be applied in various multi-objective optimisation problems. Exploring the RL-based optimisation method could be useful, as it is not directly dependent on the gradient-based approaches and learns while exploring the parameter space. 

\section*{Acknowledgements}
Funding: This work was supported by the Deutsche Forschungsgemeinschaft Priority Program: SPP 2353- Project number 501877598.

\bibliographystyle{elsarticle-num}        
\bibliography{literature}        

\end{document}